\documentclass[10pt,twocolumn,letterpaper]{article}

\usepackage[pagenumbers]{cvpr} 

\usepackage[table]{xcolor} 
\usepackage{multirow}
\usepackage{graphicx}
\usepackage{subcaption}
\newtheorem{remark}{Remark}
\usepackage{float}  
\definecolor{citecolor}{HTML}{0071BC}
\definecolor{linkcolor}{HTML}{ED1C24}
\usepackage[pagebackref=false, breaklinks=true, colorlinks, bookmarks=false,  citecolor=citecolor, linkcolor=linkcolor, urlcolor=gray]{hyperref}

\title{UAM: A Unified Attention-Mamba Backbone of Multimodal Framework for Tumor Cell Classification}

\author{Taixi Chen \qquad  Jingyun Chen \qquad Nancy Guo\thanks{Corresponding author: nguo1@binghamton.edu} \vspace{.5em}\\
State University of New York at Binghamton\\
}

\begin{document}
\maketitle

\begin{abstract}
\label{sec:ab}
Inspired by the recent success of the Mamba architecture in vision and language domains, we introduce a Unified Attention-Mamba (UAM) backbone. Unlike previous hybrid approaches that integrate Attention and Mamba modules in fixed proportions, our unified design flexibly combines their capabilities within a single cohesive architecture, eliminating the need for manual ratio tuning and improving encode capability. Building on this backbone, we further propose a multimodal UAM framework that jointly performs cell-level classification and image segmentation. Experimental results demonstrate that UAM achieves state-of-the-art performance across both tasks on public benchmarks, surpassing leading image-based foundation models. It improves cell classification accuracy from 74\% to 78\%  ($n$=349,882 cells), and tumor segmentation precision from 75\% to 80\% ($n$=406 patches).
\end{abstract}
    
\section{Introduction}
We approach the design of a backbone by examining the State-of-the-Art, the hybrid Attention-Mamba architecture, which has recently shown strong performance on image and text data. This architecture effectively combines the complementary strengths of Transformers \cite{transformer, swintransformer} and Mamba~\cite{mamba}, achieving enhanced sequence modeling capability. Jamba \cite{jamba, jamba1.5} is a representative implementation of this hybrid framework. However, its design enforces a fixed ratio between Transformer and Mamba layers, constraining architectural flexibility. Moreover, we observe that this structure tends to overfit when applied to limited image data (see Section~\ref{sec: exp} for detailed analysis). While MFuser\cite{MFuser} also adopts a hybrid configuration, it operates only as an adapter module and thus remains limited in representational capacity.

To address the lack of a dedicated backbone to overcome the inherent limitations of existing architectures, we propose UAM, a \textit{Unified Attention-Mamba} backbone. UAM introduces a highly flexible design that eliminates the need for manually tuning the ratio between Attention and Mamba layers, making it well-suited for analyzing image datasets of varying sizes and feature dimensions. Specifically, the UAM backbone is composed of two key components: the Amamba layer and the Amamba-MoE layer. The Amamba layer employs Mamba to generate context-enriched embeddings that capture long-term dependencies in linear time. These embeddings are then used as the \textit{values (V)} in a cross-attention module, while the original input embeddings serve as the \textit{queries (Q)} and \textit{keys (K)}. This design allows UAM to effectively interpret high-throughput data by seamlessly integrating Mamba-generated contextual information into the attention mechanism, enabling efficient and precise global information extraction. Building upon this, the Amamba-MoE layer concatenates the outputs from the Mamba and Attention branches, followed by a Mixture-of-Experts (MoE) module \cite{moe, moe2, moe1} to process the combined representations. Inspired by recent findings that MoE significantly enhances both Mamba and Transformer architectures \cite{moe, jamba}, this layer boosts UAM’s learning capacity and computational efficiency while maintaining strong generalization performance.

\label{sec:intro}
\begin{figure}[t]
\centering
\begin{subfigure}[b]{0.49\linewidth}
    \centering
    \includegraphics[width=\linewidth, height=33mm]{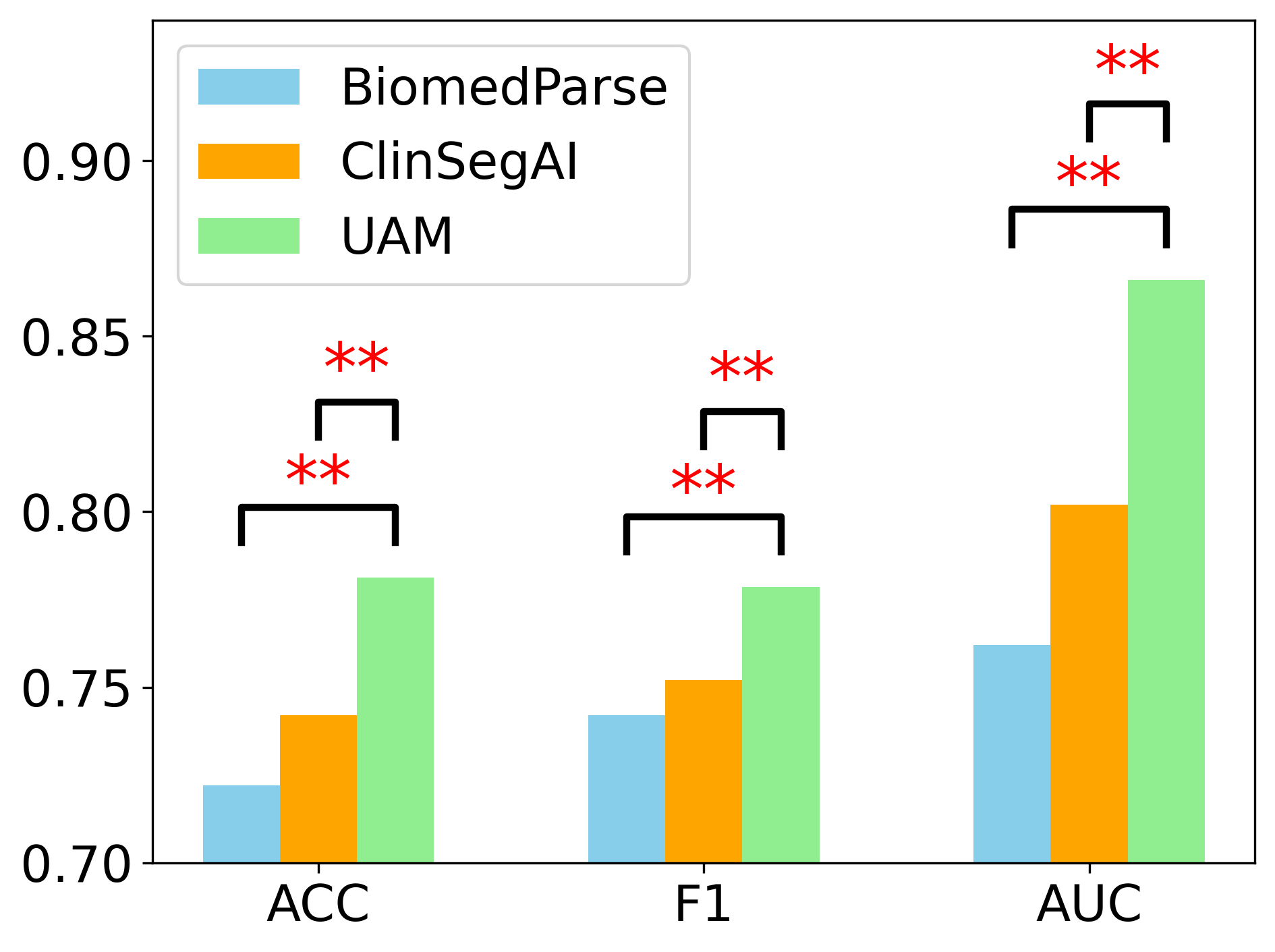}
    \caption{Results on the IGNITE dataset}
\end{subfigure}
\begin{subfigure}[b]{0.49\linewidth}
    \centering
    \includegraphics[width=\linewidth, height=33mm]{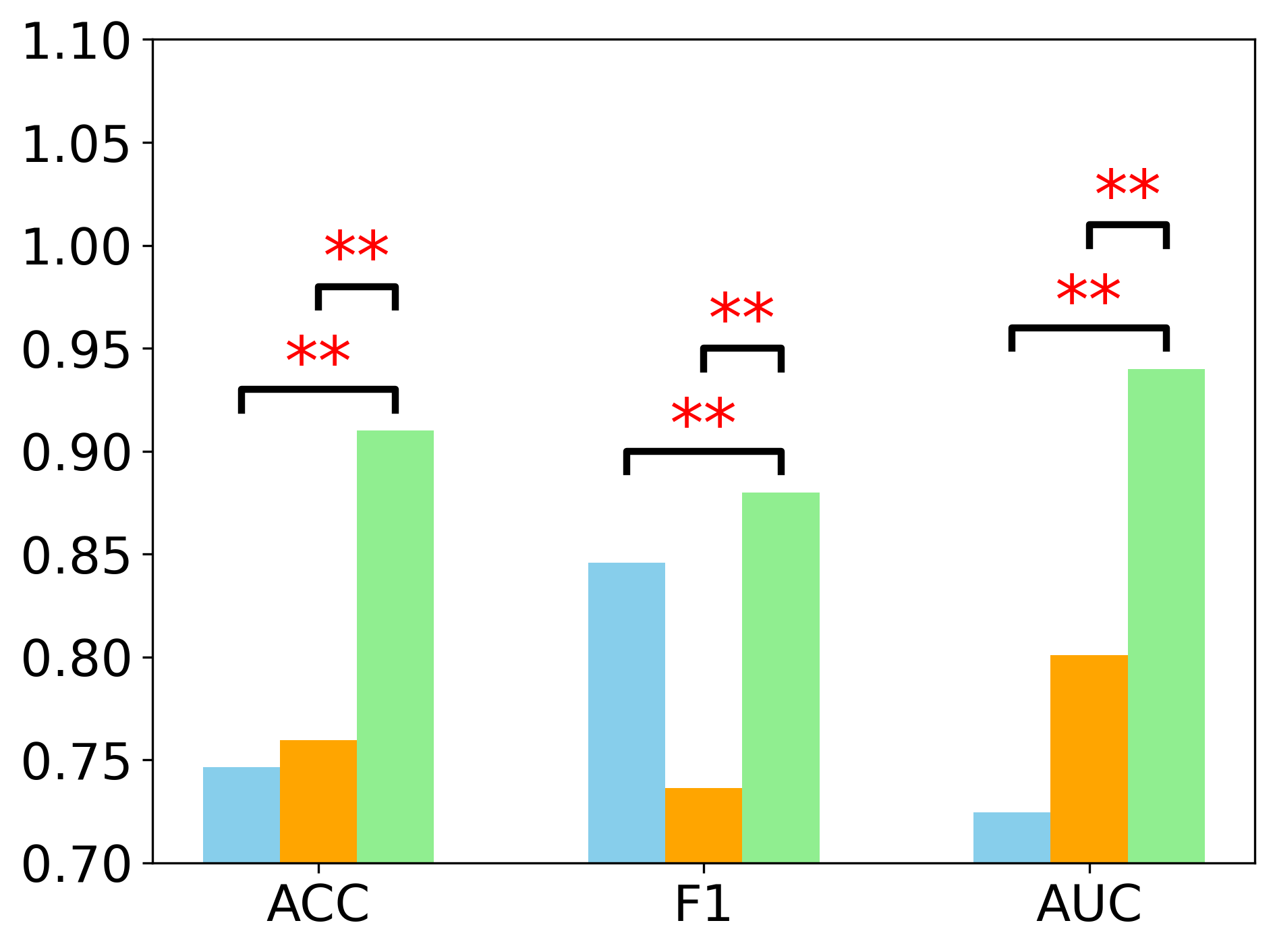}
    \caption{Results on WSSS4LUAD dataset}
\end{subfigure}
\caption{Comparison with the image-based SOTA models for cell classification on the IGNITE dataset and  WSSS4LUAD dataset. **: p-value $<$ $0.01$ (two sample t-tests). }
\label{subfig:comparison}
\end{figure}

Extending this backbone, we further introduce a multimodal UAM framework for joint cell classification and image segmentation. This framework fuses enhanced image embeddings from UAM with image embeddings from the BiomedParse encoder to produce more precise segmentation masks. Following the approach of LLaVA \cite{llava}, we project two embeddings into same space, enabling effective utilization of the pretrained BiomedParse decoder for downstream mask generation. Our contributions are summarized below: \begin{itemize}
    \item \textbf{Unified Attention-Mamba Backbone}: We present the first dedicated backbone for medical image analysis, UAM, a unified Attention-Mamba structure that flexibly models high-throughput image and text data.
    
    \item \textbf{Amamba Encoder:} We design the Amamba layer, which embeds Mamba-derived global contextual information into a cross-attention mechanism, enhancing global representation learning and model interpretability.

    \item \textbf{Amamba-MoE Encoder:} We propose the Amamba-MoE module that fuses Mamba and Attention outputs and applies a Mixture-of-Experts mechanism to further improve learning capacity and classification performance.

    \item \textbf{Multimodal Integration for Tumor Diagnosis:} We develop a multimodal UAM framework that effectively integrates enhanced image embeddings with corresponding image data. Experiments demonstrate state-of-the-art results on cell classification and image segmentation tasks, achieving 92\% cell classification accuracy and 72.06 mIoU.

\end{itemize}

\begin{figure*}
    \centering
    \includegraphics[width=1\linewidth]{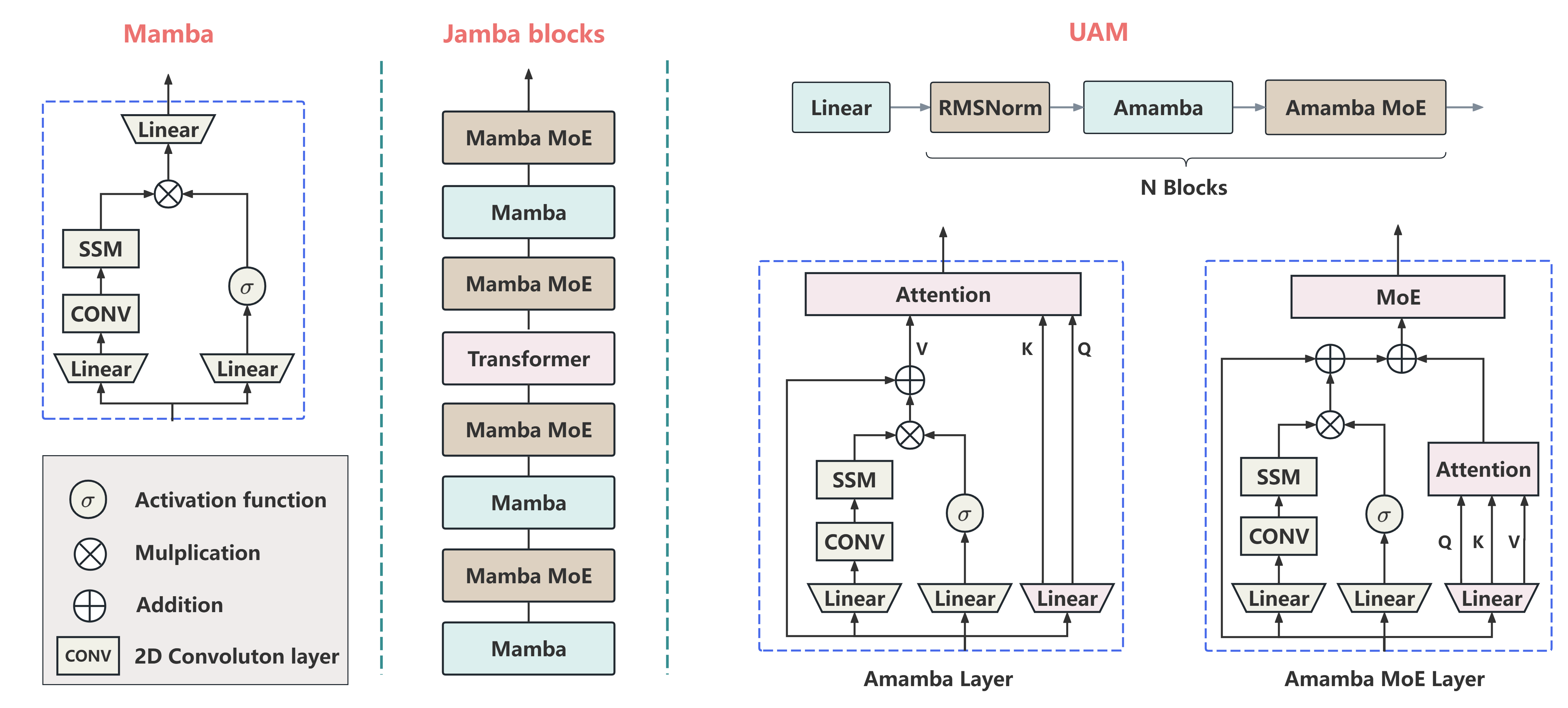}
    \caption{Overall architecture of the proposed Unified Attention-Mamba (UAM) block. Unlike Jamba, UAM integrates normalization, Amamba, and Amamba-MoE layers without fixed ratio constraints, enabling flexible fusion of attention and Mamba mechanisms. Specifically, the Amamba layer leverages Mamba to generate cross-attention values, efficiently enhancing long-range dependency modeling. Meanwhile, the Amamba-MoE layer concatenates Mamba and self-attention outputs within a mixture-of-experts (MoE) framework, providing a comprehensive, multi-perspective representation of input data for advanced processing.}
    \label{fig: main_fig}
\end{figure*}

\section{Related Work}
\label{sec:formatting}

\subsection{Hybrid Architectures}

While the Transformer architecture has long dominated both vision and language domains~\cite{dit, gpt4}, the Mamba model, built upon the State Space Model (SSM)~\cite{mamba, mamba2, tyrppg}, has recently emerged as a powerful alternative. Mamba offers linear computational complexity and excels at capturing long-range dependencies, making it increasingly attractive for large-scale sequence modeling. Recent research has extended Mamba and SSM variants to diverse domains, including GSS~\cite{mamba3}, Vision Mamba~\cite{vision-mamba}, and VM-Unet~\cite{vm-unet}. To combine the strengths of both paradigms, hybrid Attention–Mamba structures have gained growing interest. Jamba~\cite{jamba, jamba1.5} is among the first to explore this integration, identifying an optimal ratio between Transformer and Mamba layers. However, its fixed-layer ratio constrains architectural flexibility and limits generalization. MFuser~\cite{MFuser}  introduces an enhancer block to integrate Vision Foundation Models and Vision–Language Models. BMTNet~\cite{bmtnet} and Tamba \cite{Tamba} design hybrid strategies for demosaicing and forecasting tasks. Nevertheless, none of these architectures serves as a unified foundational backbone for understanding a single modality. In contrast, our proposed UAM represents the first Unified Attention–Mamba architecture, offering a flexible and efficient backbone tailored to medical image data.

\section{Methods}
\label{sec: methods}
In this section, we first formulate the medical image classification problem. We then present a detailed description of our proposed Unified Attention-Mamba model, highlighting its two key components, the Amamba and Amamba-MoE encoders. Finally, we describe the multimodal extension of UAM, which integrates enhanced and original image features for joint tumor cell classification and segmentation.

\subsection{Problem Formulation}
A medical image classification model needs to analyze the cell image $X^c$ and predict the cell labels $Y^p = \{y_i^p\}_{i=1}^N$. Due to the long sequence and well-defined values of the image features, $x^c_i$ can be expressed as $x^c_i=\{x^c_{it}\}_{t=1}^l$, where we split the image into several patches $l$. An embedding layer is initially employed for processing input $x^c_{it}$,
\begin{equation}
    x^f_{it} = \text{Linear}(x^c_{it}),
\end{equation}
where $\text{Linear(.)}$ is the linear projection. Moreover, following the Jamba~\cite{jamba}, we also add the RMSNorm~\cite{rmsnorm} at the beginning of each block, which can be expressed as:
\begin{equation}
    \bar{x}_{it}^f = \text{RMSNorm}(x_{it}^f).
\end{equation}
Further, the $\bar{x}_{it}^f$ is forwarded to the Amamba and Amamba-MoE layer, which can be defined as:
\begin{equation}
    \bar{x}_{it}^a = A^m(A(\bar{x}_{it}^f)),
\end{equation}
where $A(.)$ denotes the function of Amamba, and $A^m(.)$ denotes the Amamba-MoE layer. They are the backbone of image understanding and analysis, as shown in Figure~\ref{fig: main_fig}.

Next, we will delve into the details of the Amamba and Amamba-MoE, along with the multimodal architecture.

\subsection{Amamba Encoder}
Given the high-throughput nature of image patches, it is crucial to identify effective methods for extracting feature dependency information and generating comprehensive global representations. This paper presents a novel and practical backbone for encoding features, designed to effectively capture long-range dependency information, thereby facilitating cross-attention mechanisms to focus on important image features and generate embeddings with more meaningful global information. Given that Mamba~\cite{mamba} enables efficient extraction of long-range dependencies in linear time, we employ the SSM to encode features as the values $V_{it}$ in the cross-attention module, while using the original input embeddings as the query $Q_i$ and key $K_i$. Thus, $\bar{x}_i^f$ will first be computed as:
\begin{equation}
    \begin{aligned}
    g_{it} &= \sigma(\text{Linear}(\bar{x}_{it}^f)), \\
    h_{it} &= (1-g_{it})h_{i(t-1)} + g_{it}\bar{x}_{it}^f.
    \end{aligned}
\end{equation}
where $h_{i(t-1)}$ is the hidden state at the $t_\text{th}$ sequence derived from state space model (SSM)~\cite{mamba}. $h_{it}$ will then be embedded into cross-attention, which is defined as:
\begin{equation}
    \bar{V}_{it} = W^T_v(h_{it} +\bar{x}_{it}^f), \quad K_{it} = W^T_k\bar{x}_{it}^f, \quad Q_{it} = W^T_q\bar{x}_{it}^f,
\end{equation}
where we add one residual connection for $h_{it}$ to enhance the stability of modules, similar to the Jamba~\cite{jamba}. Thus, the final outputs of Amamba should be: 
\begin{equation}
    O_{it}^a = \text{softmax}(\frac{Q_{it}K^T_{it}}{\sqrt{d_k}})\bar{V}_{it},
\end{equation}
where $d_k$ denotes the dimensionality of the key vectors. With Amamba, UAM can encode image features into an embedding that includes richer global information. 

\subsection{Amamba-MoE Encoder}
Mixture of Experts (MoE)~\cite{moe, moe2} has gained increasing attention in both the vision and language fields, as it enables sparsely activated models with multiple expert networks while maintaining low computation costs. In this work, we incorporate the MoE mechanism into the proposed AMamba-MoE encoder. Unlike existing transformer-MoE or Mamba-MoE architectures~\cite{moe1, jamba}, our design integrates the outputs of self-attention and Mamba modules as inputs to the MoE layer, thereby enhancing model performance through hybrid structural synergy. This integration allows the model to leverage global information from multiple perspectives, enabling each expert to process richer and more diverse feature representations. Specifically, the integrated embeddings are computed by:
\begin{equation}
    \begin{aligned}
        &O_{it}^{'} = \text{Concatenate}(\text{self-atten}(O_{it}^a), \text{Mamba}(O_{it}^a)), \\
    \end{aligned}
\end{equation}
where $\text{self-atten}(.)$ represents the self-attention. The MoE then processed the integrated embeddings as:
\begin{equation}
    \begin{aligned}
        &O_{it}^{am} = \text{MoE}(O_{it}^{'}+O_{it}^a), \\
    \end{aligned}
\end{equation}
where $O^{am}_{it}$ denotes the final outputs of Amamba-MoE. Analogous to the Amamba encoder, the Amamba-MoE module incorporates a residual connection. This design aims to unify the strengths of heterogeneous architectures within a streamlined structure, thereby improving the model’s representational and learning capabilities.
\begin{remark}
    The Amamba encoder is designed to capture enhanced global information within a unified hybrid architecture for generating image embeddings. In contrast, the Amamba-MoE encoder emphasizes leveraging the fused Mamba and Attention outputs to facilitate MoE processing and model capability. Crucially, both encoders benefit from the inherent strengths of the Mamba and attention mechanisms, allowing the model to encode more effectively.
\end{remark}
Therefore, the UAM backbone leverages the Amamba and the Amamba-MoE layers to encode cell image data, facilitating cell classification. For simplicity, yet without loss of generality, the UAM is optimized by cross-entropy loss.

\subsection{Multimoal Framework}
Inspired by the success of BiomedParse~\cite{biomedparse} in medical image segmentation, we propose a multimodal framework that integrates the proposed UAM into BiomedParse, which enables the utilization of leveraging both enhanced image embeddings (From UAM), original image embeddings (from BiomedParse), and prompt text for model learning. Following the work of LLAVA~\cite{llava, llava1.5}, the embeddings from UAM will be projected into the dimensions of the image and prompt, which are expressed as:
\begin{equation}
    \begin{aligned}
        z^{r}_i &= \text{MLP}(e^{r}_i),\\
    \end{aligned}
\end{equation}
where $\text{MLP}(.)$ is a projection function, $e^r$ denotes the image embeddings from UAM. We then concatenate two embeddings as follows:
\begin{equation}
    \begin{aligned}
        z^{c}_i &= \text{Concatenate}(z^{r}_i, z^{m}_i),\\
    \end{aligned}
\end{equation}
where $z^m_i$ are the corresponding embeddings of image and prompt from the BiomedParse encoder. $z^c_i$ then can be leveraged for the BiomedParse decoder to generate tumor region mask~\cite{biomedparse}
\begin{equation}
    \begin{aligned}
        \text{mask}_i = \text{Decoder}(z^c_i).\\
    \end{aligned}
\end{equation}
Within this framework, enhanced imaged embeddings (from UAM) provide granular information about tumor cells, enhancing the representation and understanding of tumor regions, thereby enabling the model to achieve better segmentation performance. Furthermore, the proposed multimodal framework enables the utilization of different models' embeddings to identify cell-level labels, thereby expanding the applicability of BiomedParse~\cite{biomedparse} and ClinSegAI~\cite{clinsegai}. Consequently, this multimodal framework enhances the accuracy of fine-grained clinical diagnosis and supports more precise treatment planning. During the training stage, we utilize the original loss of BiomedParse~\cite{biomedparse} plus the classification loss to optimize the whole model.

\section{Experiments}
\label{sec: exp}
In this section, we first describe the experimental settings and implementation details. Next, we present the results of the proposed UAM and its multimodal extension, comparing them against several baseline methods. Finally, we provide visualizations of cell classification results of UAM.
\subsection{Experiment Settings}
\begin{table*}[t]
\centering
\small
  \caption{Tumor cell classification results in comparison with baselines on four datasets. \textit{Combined}: WSSS4LUAD + IGNITE; \textit{+TCGA}: WSSS4LUAD + IGNITE + TCGA.  \colorbox{green!20}{\textcolor{black}{Green}} highlights: UAM variants. \textbf{Bold}: best performance.}
    \begin{tabular}{c |c| c c c c c| >{\columncolor{green!20}}c|>{\columncolor{green!20}}c|>{\columncolor{green!20}}c}
    \toprule
      Datasets & Metric  & Trans~\cite{transformer} & Trans-M~\cite{moe1} & Mamba~\cite{mamba} & Mamba-M~\cite{jamba} & Jamba~\cite{jamba} & UAM-L & UAM-M & UAM \\
      \midrule
        \multirow{3}{*}{WSSS4LUAD} &  Accuracy & 85.77\% & 86.52\% & 90.65\% &  89.35\%& 88.96\%  & 91.71\% &90.41\% &\textbf{92.06\%}\\
        &  F1  & 81.53\% & 81.71\% & 88.27\% & 87.02\%& 85.31\%  & 88.72\%& 87.90\% & \textbf{89.70\%}\\
        &  AUC   & 91.81\% & 91.73\% & 94.60\%& 95.67\%  &94.63\% &93.40\%& 93.80\% & \textbf{95.90\%}\\
        \midrule
        \multirow{3}{*}{IGNITE} &  Accuracy   & 77.56\% & 77.83\% & 77.95\% & 77.94\% & 78.05\% & 77.97\% & 77.87\% & \textbf{78.53\%}\\
        &  F1  & 76.88\% & 76.89\% & 77.49\%& 77.46\% & 77.23\% & 77.55\%& 77.59\% & \textbf{77.73\%}\\
        &  AUC  & 86.13\% & 86.40\%&  86.60\% & 86.52\% & 86.12\%& 86.62\% & 86.55\%&\textbf{86.71\%}\\
        \midrule
        \multirow{3}{*}{Combined} & Accuracy  & 78.71\%& 78.64\% & 78.40\% & 79.50\% &  79.45\% & 79.55\% & 79.39\%& \textbf{80.70\%}\\
        &  F1  & 77.11\% & 76.97\% & 76.99\% & 77.94\% & 77.69\%& 77.96\% &78.28\% &\textbf{79.34\%}\\
        &  AUC  & 87.05\% & 87.13\% & 86.82\% & 87.64\%& 87.80\%&  87.87\%& 87.91\% & \textbf{88.80\%}\\
        \midrule
        \multirow{3}{*}{+TCGA} &  Accuracy   & 82.27\% & 82.70\% & 83.01\% & 83.33\%& 78.61\%& 84.20\%& 84.24\% & \textbf{84.33\%}\\
        &  F1   & 79.70\%& 79.79\% & 80.57\%& 80.57\% &71.69\%&81.48\%& \textbf{81.53\%} &81.45\%\\
        &  AUC  & 90.37\% & 90.50\% & 91.18\% & 91.15\% &75.60\%&91.79\%&\textbf{91.94\%} &91.91\%\\
        \bottomrule
    \end{tabular}
  \label{tab:initial-cls}
\end{table*}

\begin{table*}[t]
\centering
\small
  \caption{Cross-validation results for tumor cell classification with competitive baselines. Models are trained on the IGNITE dataset and tested on other datasets. \textit{+TCGA}: WSSS4LUAD + TCGA. \colorbox{green!20}{\textcolor{black}{Green}} highlights: UAM variants. \textbf{Bold}: best performance.}
    \begin{tabular}{c|c| c c c c c |>{\columncolor{green!20}}c|>{\columncolor{green!20}}c|>{\columncolor{green!20}}c }
    \toprule
      Datasets & Metric&  Trans~\cite{transformer} & Trans-M~\cite{moe1} & Mamba~\cite{mamba} & Mamba-M~\cite{jamba}& Jamba~\cite{jamba} & UAM-L & UAM-M & UAM \\
      \midrule
        \multirow{3}{*}{WSSS4LUAD} &  Accuracy& 77.54\% &  79.70\% &77.58\% &  78.45\% & 76.62\% & 78.96\% &79.27\% &\textbf{81.76\%}\\
        &  F1 & 73.45\% & 73.42\% &70.26\% & 71.65\%& 70.21\%&73.49\%& 72.36\% & \textbf{74.84\%}\\
        &  AUC & 83.56\% & 83.48\% & 76.14\% & 82.13\% & 80.63\% &\textbf{84.75\%}& 83.80\% & 84.46\%\\
        \midrule
        \multirow{3}{*}{+TCGA} &  Accuracy& 80.65\% & 77.29\%& 66.48\%  & 64.65\% & 56.37\% & \textbf{88.22\%} &72.27\% &86.57\%\\
        &  F1 & 80.13\% & 65.73\% &62.55\% & 65.36\% & 50.68\%&\textbf{82.38\%}& 70.25\% & 81.06\%\\
        &  AUC & 91.45\% & 73.51\% & 64.34\% & 68.62\% & 57.63\% &96.13\%& 66.80\% & \textbf{98.54\%}\\
        
        \bottomrule
    \end{tabular}
  \label{tab:cross-cls}
\end{table*}


\begin{figure*}[htbp]
	\centering
	\begin{subfigure}{0.22\linewidth}
		\centering
		\includegraphics[width=38mm, height=46mm]{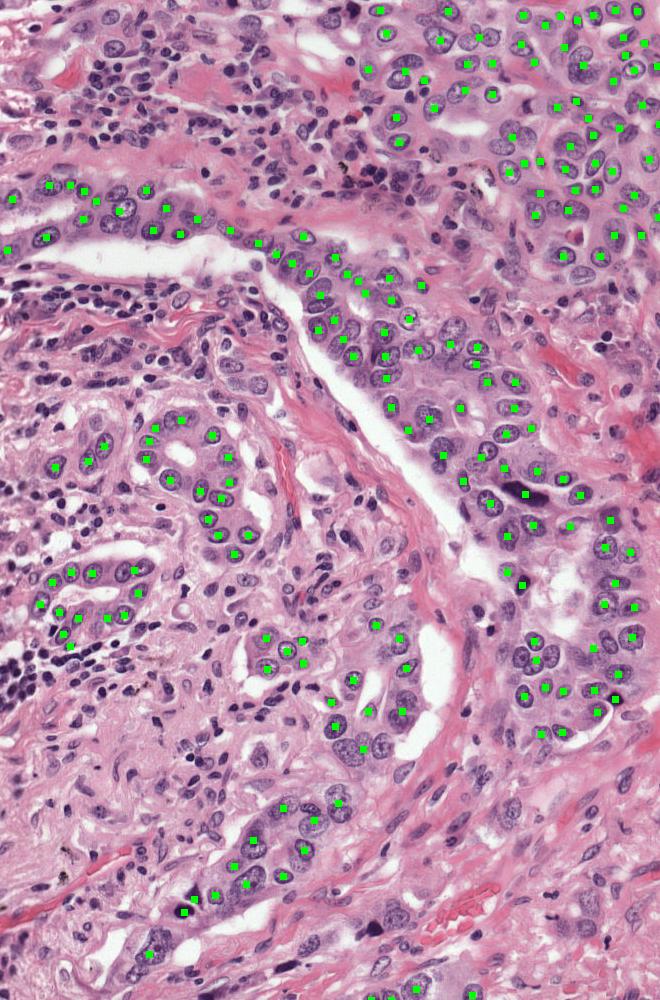}
		\caption{Ground Truth1}
		
	\end{subfigure}
	\centering
	\begin{subfigure}{0.22\linewidth}
		\centering
		\includegraphics[width=38mm, height=46mm]{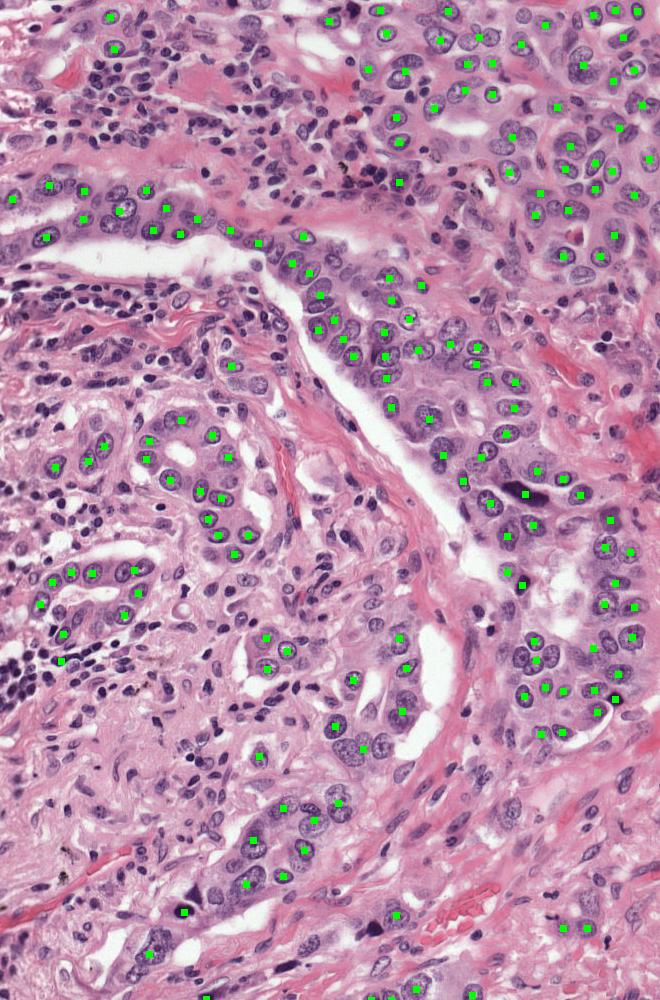}
		\caption{Prediction1}
		
	\end{subfigure}
	\centering
	\begin{subfigure}{0.22\linewidth}
		\centering
		\includegraphics[width=38mm, height=46mm]{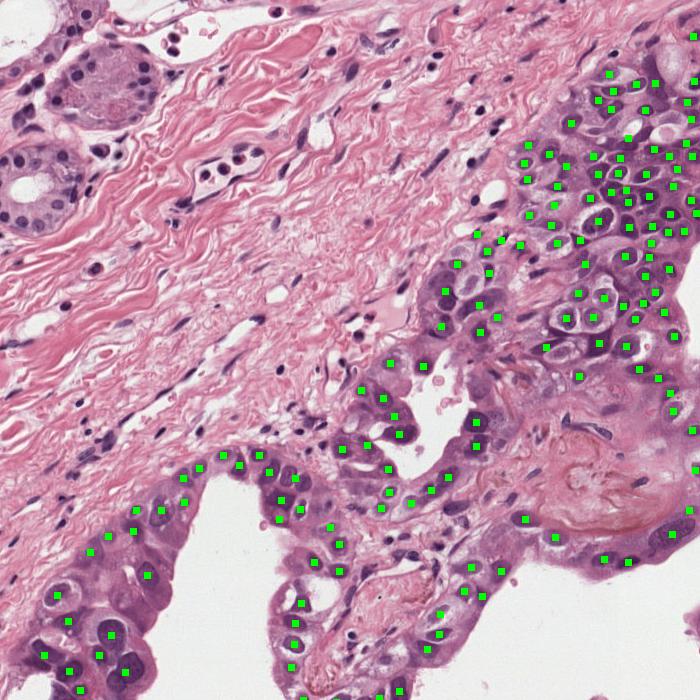}
		\caption{Ground Truth2}
		
	\end{subfigure}
    \begin{subfigure}{0.22\linewidth}
		\centering
		\includegraphics[width=38mm, height=46mm]{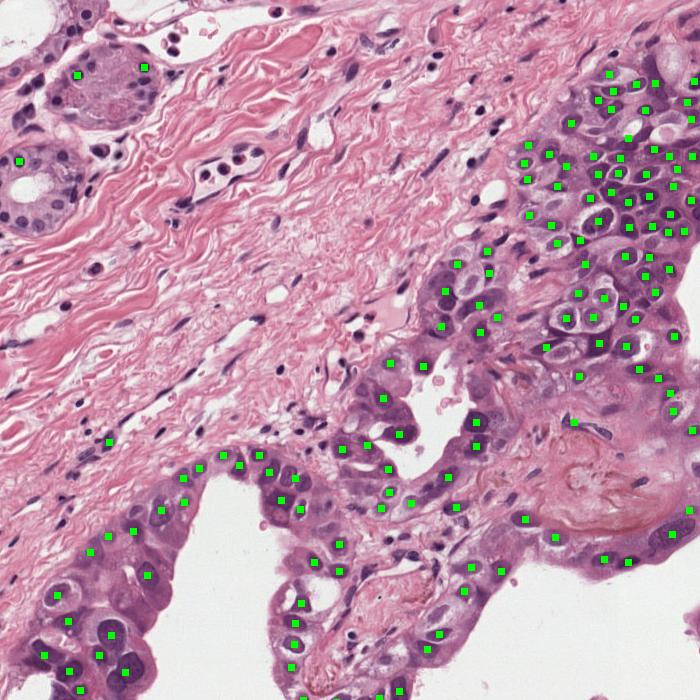}
		\caption{Prediction2}
		
	\end{subfigure}
	\caption{Visual comparison of ground truth and UAM predictions on the IGNITE dataset. Tumor cells are highlighted in green.}
	\label{subfig:vis}
\end{figure*}

\begin{table*}[t]
\centering
  \caption{Comparative image segmentation results between the proposed multimodal model and BiomedParse. \textit{Combined+TCGA}: WSSS4LUAD + IGNITE + TCGA\textit{.} \textbf{Bold}: best performance. Symbol $\uparrow$: higher values are preferred.}
    \begin{tabular}{c|c |c c c c c}
    \toprule
      Datasets & Methods & Precision $\uparrow$  &cIoU $\uparrow$ & mIoU $\uparrow$ & cDICE $\uparrow$ & mDICE $\uparrow$\\
      \midrule
        \multirow{2}{*}{WSSS4LUAD}
        & BiomedParse & 88.04 & 80.16 & 73.48 &  88.98 & 83.52\\
        & UAM & \textbf{90.21} &\textbf{80.64} & \textbf{73.69} & \textbf{89.28} & \textbf{83.45}\\
        \midrule
        \multirow{2}{*}{IGNITE}
        & BiomedParse & 75.34 & 74.13 & 63.47 & 85.14 & 73.51\\
        & UAM &  \textbf{80.82} & \textbf{74.94} & \textbf{66.71} & \textbf{85.67} & \textbf{76.83}\\
        \midrule
        \multirow{2}{*}{Combined+TCGA}
        & BiomedParse & 88.54 & 76.95 & 70.86 &86.97 & 81.05\\
        & UAM & \textbf{90.45}& \textbf{78.11} & \textbf{72.06} & \textbf{87.71} & \textbf{81.85}\\
        \bottomrule
    \end{tabular}
  \label{tab:seg}
\end{table*}

\begin{table}[t]
\centering
\small
  \caption{Efficiency comparison with competitive baselines. Symbol $\downarrow$: lower values are preferred.}
    \begin{tabular}{c| c c }
    \toprule
      Methods &  Flops (K) $\downarrow$ & Parameters (K) $\downarrow$\\
      \midrule
         Trans~\cite{transformer} & 384.26 & 6.642     \\   
         
         Mamba~\cite{mamba} & 305.41 & 5.546   \\ 
         Jamba~\cite{jamba} & 470.02 & 15.858 \\
         \midrule
         UAM-L & 436.48& 7.722\\
         UAM-M  & 379.14& 6.770 \\
         UAM   & 407.81 & 13.554\\
         
        \bottomrule
    \end{tabular}
  \label{tab:flops}
\end{table}

\textbf{Datasets.} We evaluated our model based on three cancer datasets, including WSSS4LUAD~\cite{wsss4luad}, IGNITE~\cite{IGNITE}, and TCGA~\cite{TCGA_LUAD}. The WSSS4LUAD dataset includes $309$ H\&E patches and corresponding annotated masks (i.e., tumor or non-tumor), with a resolution of $1024 \times 1024$. The IGNITE dataset includes $406$ H\&E patches with the same resolutions, and its annotations are generated from both machine and experts. The TCGA dataset contains 848 selected H\&E normal tissue patches to balance the distribution of class labels in WSSSLUAD and IGNITE datasets. Specifically, the WSSS4LUAD dataset contained 153,702 cells, the IGNITE dataset included 349,882 cells, and the TCGA dataset comprised 112,001 cells. Except for cross-validation, we divided the four datasets into training and testing sets using an 8:2 ratio, ensuring the split is performed based on distinct individuals.

\textbf{Implementation Details.}
During training, the number of UAM blocks and other baseline modules was set to 4, with additional details provided in the ablation studies. The batch size was set to 4. The AdamW optimizer~\cite{adamw} was employed with a learning rate $1\times 10^{-4}$. All experiments were conducted on a single  NVIDIA RTX A6000 GPU with 49 GB of memory.

\subsection{Comparison with State-of-The-Art Methods}
\textbf{Tumor Cell Classification.}
We first evaluated the model performance on the cell classification task using only UAM. Since most vision and language models are not capable of handling this setting, we replaced the main block of the UAM with several state-of-the-art (SOTA) architectures for comparison, including Transformer~\cite{transformer} (abbreviated as Trans), Transformer-MoE~\cite{moe1} (abbreviated as Trans-M), Mamba~\cite{mamba}, Mamba-MoE~\cite{jamba} (proposed in Jamba and abbreviated as Mamba-M), and Jamba~\cite{jamba}. In addition, we introduced two ablated variants of UAM in comparison: UAM-L, which includes only the Amamba encoder, and UAM-M, which contains only the Amamba-MoE encoder. These variants serve as an ablation study of the UAM architecture, further discussed in Section~\ref{subsec:ablation}. 

As shown in Table~\ref{tab:initial-cls}, our proposed methods consistently outperform all baselines across three metrics, accuracy, F1 score, and AUC. Specifically, UAM achieves 92.06\% accuracy on the WSSS4LUAD dataset, surpassing Mamba (90.65\%), Mamba-M (89.35\%), Jamba (88.96\%), and Transformer (85.77\%). Moreover, both UAM variants outperform competing methods on the IGNITE dataset as well as on the combined dataset. These results demonstrate that UAM effectively enhances global information extraction through its unified architecture, validating the efficacy of the proposed approach.

Furthermore, we evaluated the generalization ability of UAM in comparison with several SOTA baselines, as shown in Table~\ref{tab:cross-cls}. Specifically, each model was trained on the IGNITE dataset, which contains the largest number of cell samples, and subsequently tested on the WSSS4LUAD dataset as well as on the combined WSSS4LUAD–TCGA dataset. UAM consistently outperforms all competing baselines, demonstrating its superior capability in learning universal representations.

Notably, Jamba exhibited severe generalization deficiencies using the same number of blocks as other baselines. This observation suggests that its rigid hybrid design, which enforces a fixed ratio between Transformer and Mamba components, is suboptimal for medical image analysis and prone to overfitting.

\textbf{Comparisons with Imaged-based SOTA}. We also compared UAM with two image-based SOTA methods, including BiomedParse~\cite{biomedparse} and ClinSegAI~\cite{clinsegai}. As shown in Figure~\ref{subfig:comparison}, UAM achieves significantly higher accuracy than both BiomedParse and ClinSegAI (p$<$0.01, two-sample t-tests) on the IGNITE and WSSS4LUAD datasets. These results demonstrate that our UAM effectively exploits richer cellular feature information and surpasses image-based approaches, thereby improving diagnostic accuracy. More details about their cell classification process are illustrated in the \textit{Supplemental Material}.

\begin{figure}[t]
\centering
\begin{subfigure}[b]{0.49\linewidth}
    \centering
    \includegraphics[width=\linewidth, height=35mm]{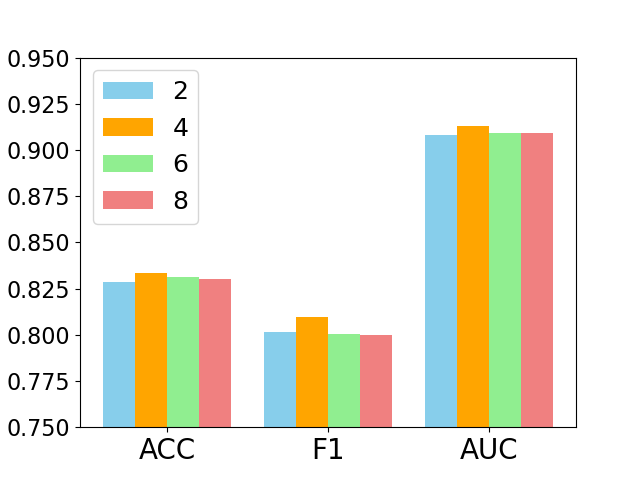}
    \caption{Results on the Combined dataset}
\end{subfigure}
\begin{subfigure}[b]{0.49\linewidth}
    \centering
    \includegraphics[width=\linewidth, height=35mm]{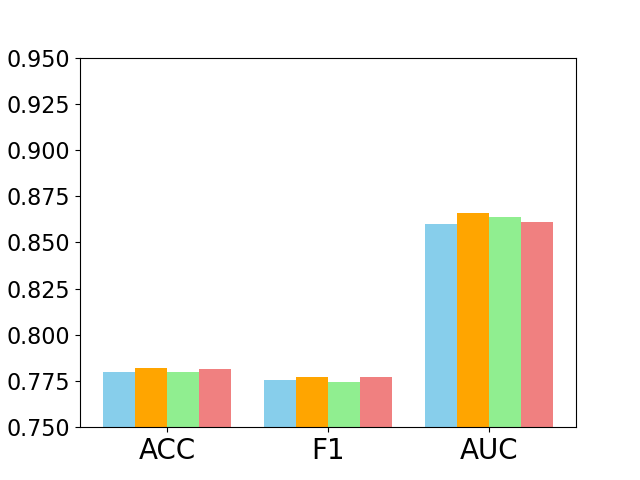}
    \caption{Results on the IGNITE dataset}
\end{subfigure}
\caption{Ablation studies evaluating the impact of varying the number of UAM blocks on the IGNITE dataset and the combined dataset (WSSS4LUAD + IGNITE). }
\label{subfig:ablations}
\end{figure}

\textbf{Tumor Segmentation.} We further evaluated the proposed multimodal UAM framework on the image segmentation task. The Precision, cIoU, mIoU, cDICE, and mDICE metrics are used to evaluate the segmentation performance. For more evaluation details, please refer to the \textit{Supplemental Material}. As shown in Table~\ref{tab:seg}, integrating UAM notably enhances the segmentation performance of BiomedParse. Specifically, the multimodal model improves the precision from $75.34$ to $80.02$ on the IGNITE dataset, and increases mIOU from $70.86$ to $72.06$ on the combined dataset. These results highlight the effectiveness of embeddings generated by UAM and demonstrate that incorporating fine-grained cell information substantially improves tumor segmentation. This underscores the broad applicability of UAM for enhancing multimodal medical analysis. 

\textbf{Efficiency.} We evaluated the computational efficiency of UAM by measuring its Flops and total number of parameters in comparison with SOTA models, as shown in Table~\ref{tab:flops}. As a unified Attention-Mamba structure, UAM naturally exhibits higher FLOPs and parameter counts than models based solely on Transformers or Mamba. However, it achieves significantly lower FLOPs and parameter complexity than Jamba, demonstrating UAM’s superior efficiency and flexibility resulting from its unified design. Notably, UAM-M achieves even lower FLOPs than Transformer, indicating that the Amamba-MoE encoder enables faster processing through its streamlined hybrid structure.

\textbf{Visualization and AI Interpretability.} As illustrated in Figure~\ref{subfig:vis},  UAM effectively highlights tumor cells on H\&E slides based on image data, facilitating pathologist interpretation and enhancing AI explainability.

\subsection{Ablation Study}
\label{subsec:ablation}

We conducted ablation experiments with two model variants, UAM-L and UAM-M, to systematically evaluate the contributions of different components. As shown in Table~\ref{tab:initial-cls}, UAM-M achieves the best overall performance when all datasets are combined, demonstrating the effectiveness of the Amamba-MoE encoder. This confirms our hypothesis that the Mixture-of-Experts (MoE) mechanism enhances the image representation capability of hybrid architectures, particularly on large-scale datasets. Furthermore, Table~\ref{tab:cross-cls} shows that UAM-L exhibits comparable generalization performance to the full UAM model, indicating that the Amamba encoder effectively captures universal image representations. Nevertheless, UAM consistently outperforms both ablated variants across most evaluation scenarios, underscoring the complementary advantages of combining the Amamba and Amamba-MoE encoders.

In addition, we varied the number of UAM blocks to assess the impact of this hyperparameter (set to Block = 4 in our final configuration). As illustrated in Figure~\ref{subfig:ablations}, increasing the number of blocks leads to overfitting in the limited data size, whereas reducing the block number to two causes a performance drop.

\section{Conclusion}
\label{sec: conclusion}
In this work, we presented UAM, the first dedicated backbone for cell image analysis and tumor classification. UAM unifies the strengths of the Transformer and Mamba architectures within a single framework, eliminating the need for ratio tuning and substantially enhancing image representation learning. It comprises two core components: the Amamba encoder, which leverages Mamba-generated embeddings within a cross-attention mechanism to capture global dependencies, and the Amamba-MoE encoder, which integrates a Mixture-of-Experts module to efficiently process fused attention-Mamba outputs and improve model capacity. Moreover, we introduced a multimodal UAM framework that integrates image embeddings from UAM and BiomedParse for joint tumor classification and segmentation. Extensive experiments demonstrate that UAM achieves state-of-the-art performance across multiple datasets and tasks, validating its effectiveness, flexibility, and generalization ability. These results highlight the potential of UAM as a foundation for cancer diagnosis and the integrative analysis of multimodal biomedical data.

{
    \small
    \bibliographystyle{ieeenat_fullname}
    \bibliography{main}
}

\clearpage
\appendix
\section{Cell Radiomics Dataset Preparation}
Cell-level radiomic features were extracted from H\&E-stained images using cell masks generated by ClinSegAI \cite{clinsegai}. Feature computation was performed using the PyRadiomics package (v3.1.0) in Python. The features encompassed shape-based metrics (e.g., area, perimeter), first-order intensity statistics (e.g., mean, standard deviation), and texture descriptors derived from Gray-Level Co-occurrence Matrices (GLCM) and related filters. Summary statistics for the three datasets mentioned in the article are presented in Table \ref{table1} (i.e., WSSS4LUAD dataset is abbreviated as WSSS). Specifically, the WSSS dataset~\cite{wsss4luad} focuses on Lung Adenocarcinoma (LUAD) and originally categorizes cells into three distinct classes: tumor, stroma, and normal. In stark contrast, the IGNITE dataset~\cite{IGNITE} is significantly richer, encompassing fifteen different cell types (e.g., tumor, inflammatory, and liver cells) and covers a broader range of Non-Small Cell Lung Cancer (NSCLC) subtypes, including Adenocarcinoma (AD), Squamous Cell Carcinoma (SC), and Large Cell (LC). Moreover, we selected only the normal tissue slides from the TCGA cohort~\cite{TCGA_LUAD} to balance the distribution of class labels in the WSSS and IGNITE datasets.

\begin{table}[t]
    \centering
    \caption{Statistics of three datasets.}
        \begin{tabular}{c|c|c|c|c}
            \toprule
            Dataset & Total Cells & Tumor cells & Attribute  & Types \\
             \midrule
            WSSS & 153,702 & 41,321 &106 & 3\\
            \midrule
            IGNITE & 349,882 &  151,122   &106 & 15\\
            \midrule
            TCGA & 112,001 &  0  &106 & 1\\
             \bottomrule
        \end{tabular}
    \label{table1}
\end{table}

The ground truth for each cell's true class was established by comparing regional ground truth masks (created by human experts, containing both cellular and non-cellular areas) against the ClinSegAI cell-level masks (generated by BiomedParse~\cite{biomedparse} using the prompt "cells"). This rigorous comparison process ensured the reliability of the cell-level ground truth labels.

\section{Image-based Segmentation Model}
In this study, initial tumor segmentations were generated for each slide image using BiomedParse with the prompt "Neoplastic Cells." Cell-level tumor prediction was subsequently derived by calculating the overlap between each cell mask and the resulting tumor segmentation. Consistent with UAM's (Unified Attention-Mamba) configuration, a cell was classified as tumor if its tumor coverage exceeded $50\%$, and non-tumor otherwise.

The ClinSegAI framework, which refines $\text{BiomedParse}$ segmentations through multi-scale post-processing algorithms \cite{clinsegai}, was also utilized for comparison. To generate ClinSegAI-based tumor predictions, we followed the identical procedure: generating the initial tumor segmentations (BiomedParse with the prompt "neoplastic cells"), and then applying the same $50\%$ threshold to calculate per-cell tumor coverage for classification.

Notably, the cell-level ground truth used for evaluation (which incorporates expert input and a different $\text{ClinSegAI}$ prompt) is distinct from the cell class predictions (tumor or non-tumor) generated solely by the ClinSegAI method.

\section{Evaluation Metric}
In the image segmentation task, we utilize the precision, cIoU, mIoU, cDICE, and mDICE. All metrics utilize a segmentation threshold of 0.5. Specifically, cIoU (Class IoU) represents the Intersection over Union (IoU) score calculated only for the tumor class. mIoU is the average IoU score  across all $n$ classes, which is calculated as:
\begin{equation}
\text{mIoU} = \frac{1}{n}\sum_i^n\text{IoU}_i
\end{equation}
The cDICE and mDICE scores are calculated analogously: cDICE focuses solely on the tumor class segmentation performance, while mDICE computes the average DICE score across all classes.

\begin{figure}[h]
	\centering
	\begin{subfigure}{0.31\linewidth}
		\centering
		\includegraphics[width=25mm, height=37mm]{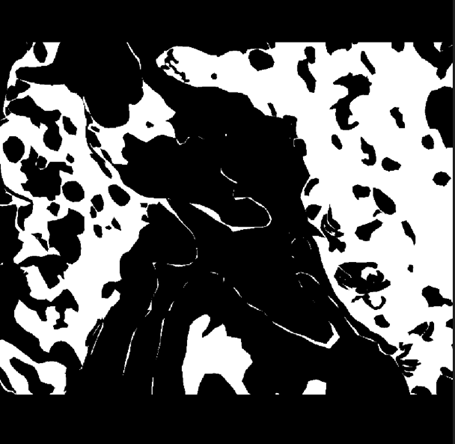}
		\caption{Ground Truth}
		
	\end{subfigure}
	\centering
	\begin{subfigure}{0.31\linewidth}
		\centering
		\includegraphics[width=25mm, height=37mm]{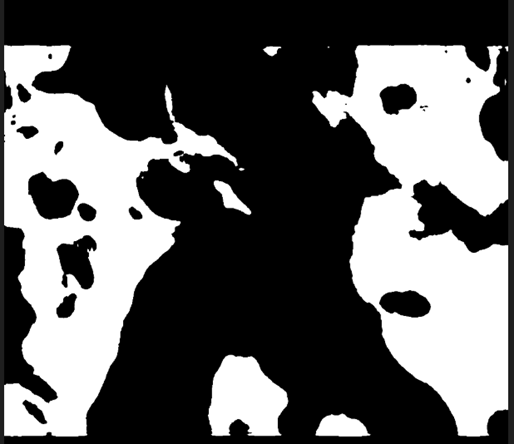}
		\caption{BiomedParse}
		
	\end{subfigure}
	\centering
	\begin{subfigure}{0.31\linewidth}
		\centering
		\includegraphics[width=25mm, height=37mm]{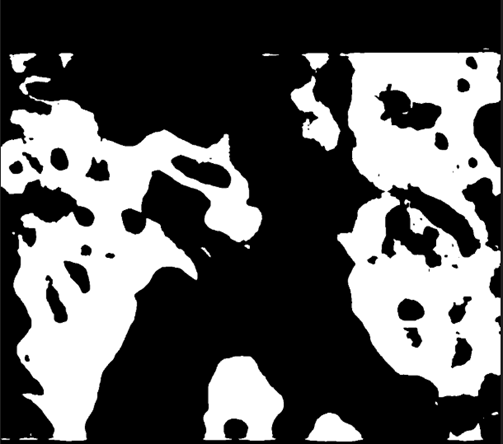}
		\caption{Multimodal UAM}
		
	\end{subfigure}
	\caption{Visual comparison of ground truth, BiomedParse, and UAM predictions on the IGNITE dataset. The masks indicate the tumor regions.}
	\label{subfig:vis2}
\end{figure}

\section{More Experiment Results}
\begin{figure*}[ht]
\vspace{-14pt}
	\centering
	\begin{subfigure}{0.22\linewidth}
		\centering
		\includegraphics[width=38mm, height=46mm]{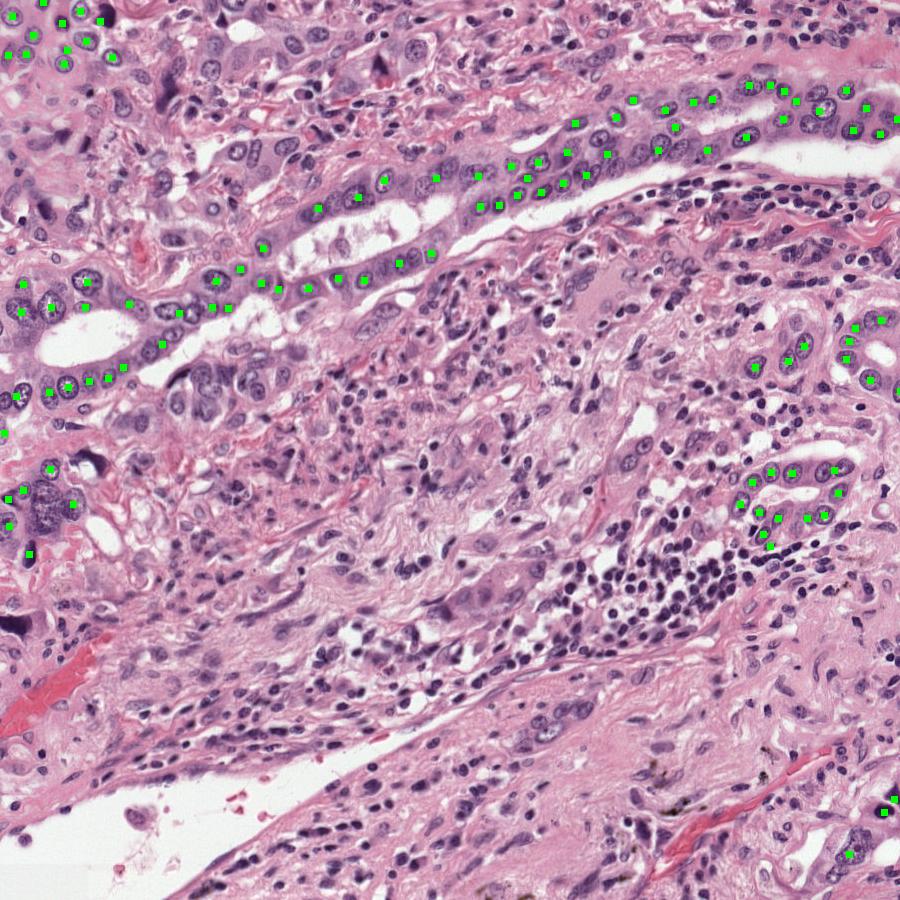}
		\caption{Ground Truth1}
		
	\end{subfigure}
	\centering
	\begin{subfigure}{0.22\linewidth}
		\centering
		\includegraphics[width=38mm, height=46mm]{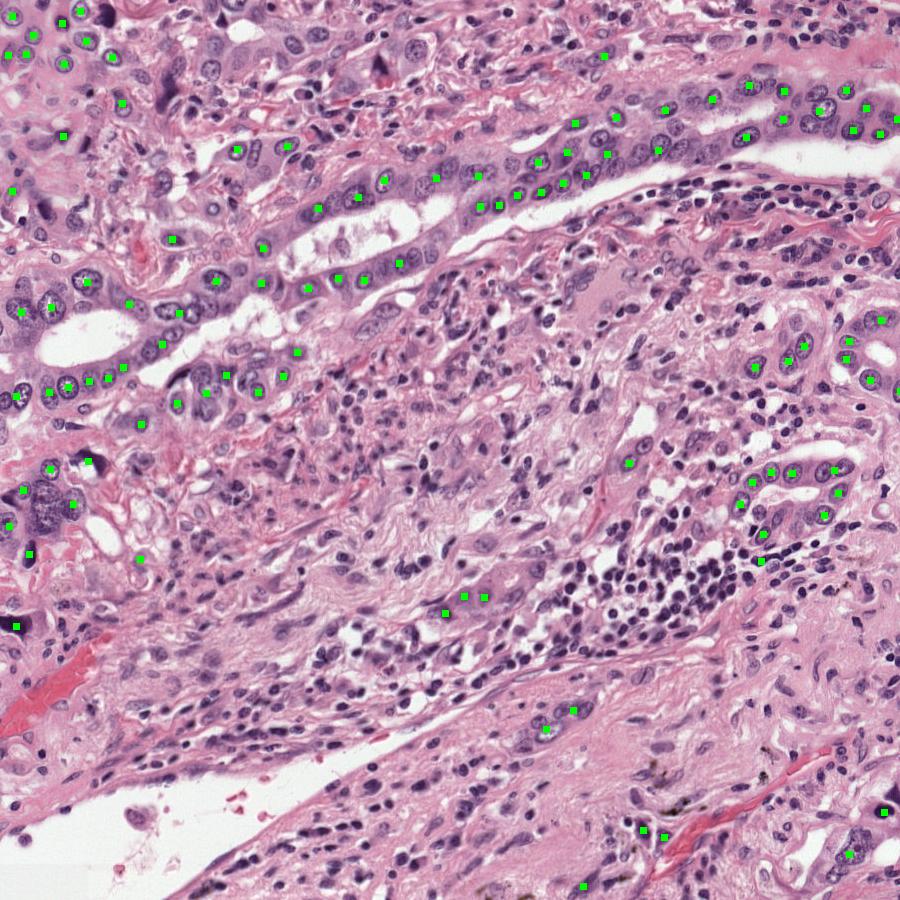}
		\caption{Prediction1}
		
	\end{subfigure}
	\centering
	\begin{subfigure}{0.22\linewidth}
		\centering
		\includegraphics[width=38mm, height=46mm]{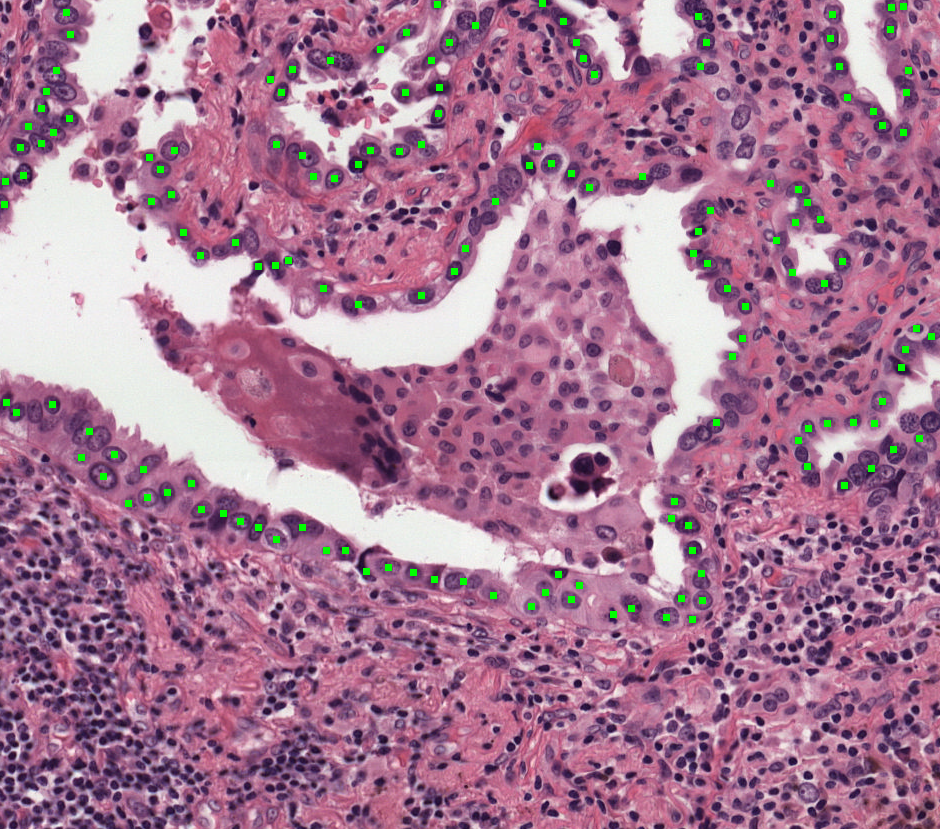}
		\caption{Ground Truth2}
		
	\end{subfigure}
    \begin{subfigure}{0.22\linewidth}
		\centering
		\includegraphics[width=38mm, height=46mm]{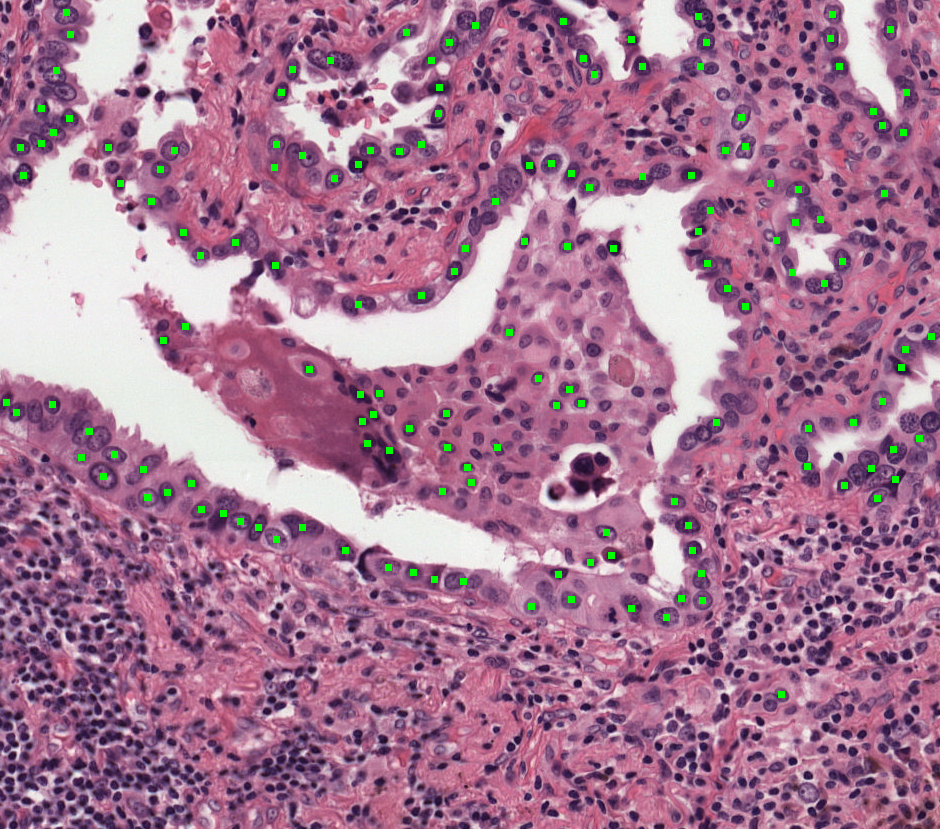}
		\caption{Prediction2}
		
	\end{subfigure}
    \vspace{-5pt}
	\caption{Visual comparison of ground truth and UAM predictions. Tumor cells are highlighted in green.}
	\label{subfig:vis3}
\end{figure*}

\begin{figure*}[htbp]
\vspace{-9pt}
	\centering
	\begin{subfigure}{0.22\linewidth}
		\centering
		\includegraphics[width=38mm, height=46mm]{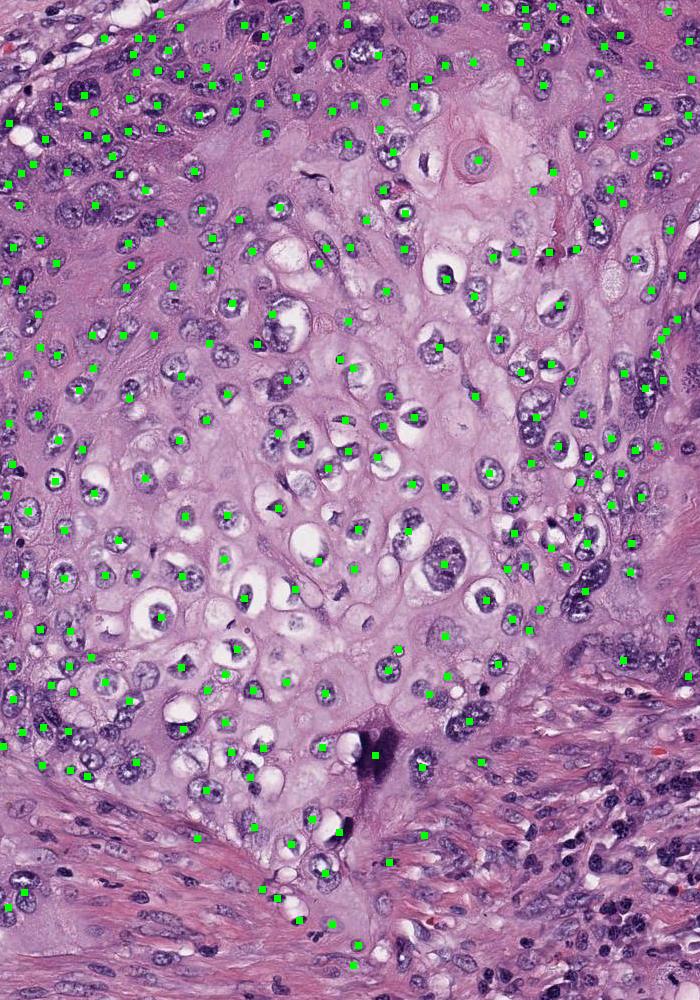}
		\caption{Ground Truth1}
		
	\end{subfigure}
	\centering
	\begin{subfigure}{0.22\linewidth}
		\centering
		\includegraphics[width=38mm, height=46mm]{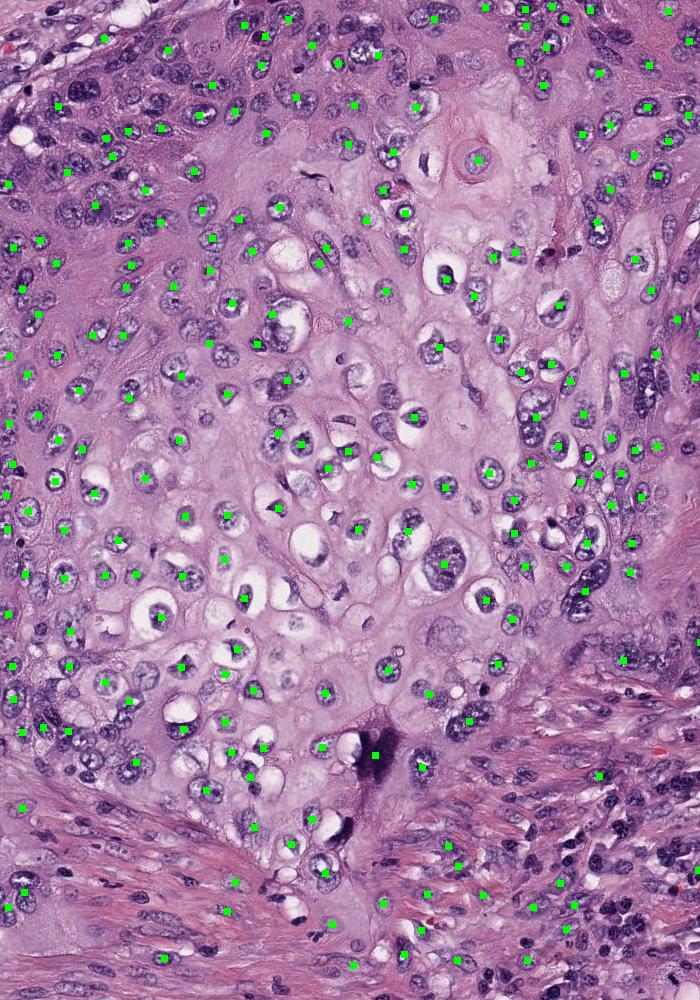}
		\caption{Prediction1}
		
	\end{subfigure}
	\centering
	\begin{subfigure}{0.22\linewidth}
		\centering
		\includegraphics[width=38mm, height=46mm]{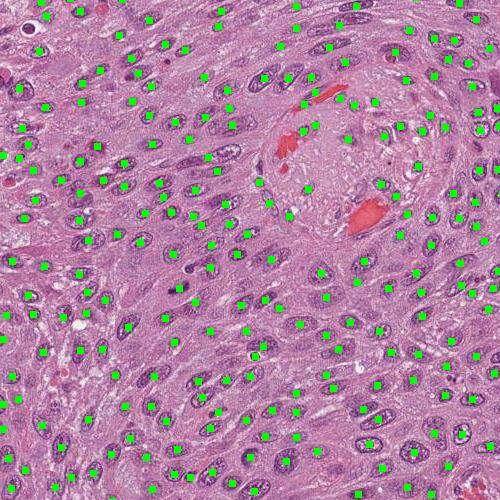}
		\caption{Ground Truth2}
		
	\end{subfigure}
    \begin{subfigure}{0.22\linewidth}
		\centering
		\includegraphics[width=38mm, height=46mm]{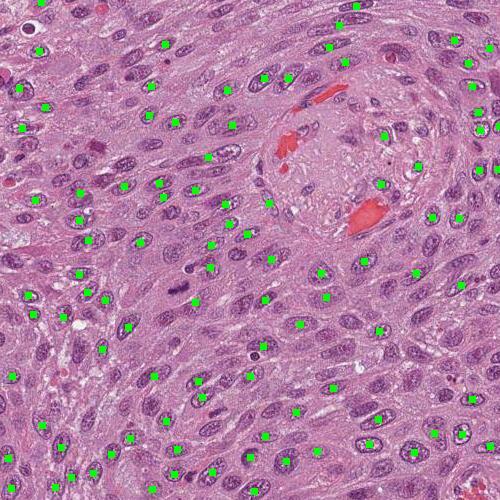}
		\caption{Prediction2}
		
	\end{subfigure}
    \vspace{-5pt}
	\caption{Visual comparison of ground truth and UAM predictions. Tumor cells are highlighted in green.}
	\label{subfig:vis4}
\end{figure*}

\begin{figure*}[htbp]
\vspace{-9pt}
	\centering
	\begin{subfigure}{0.22\linewidth}
		\centering
		\includegraphics[width=38mm, height=46mm]{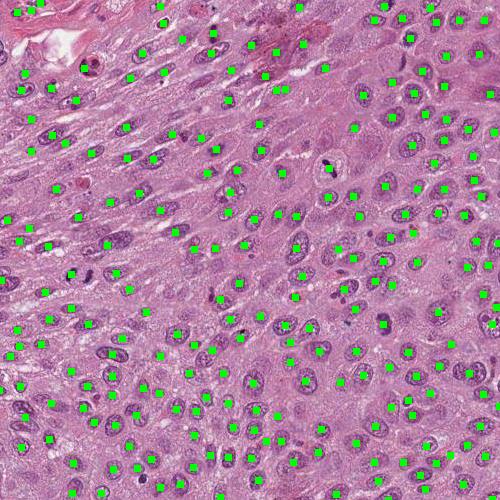}
		\caption{Ground Truth1}
		
	\end{subfigure}
	\centering
	\begin{subfigure}{0.22\linewidth}
		\centering
		\includegraphics[width=38mm, height=46mm]{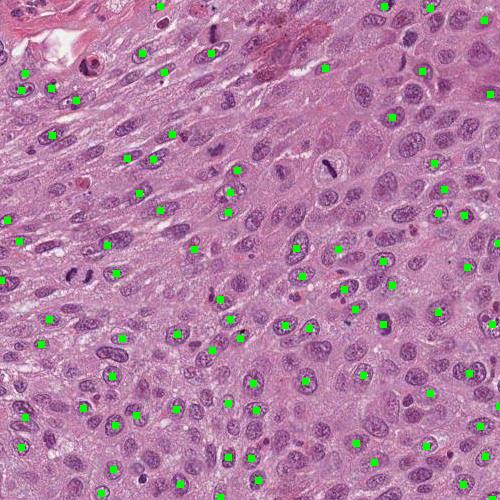}
		\caption{Prediction1}
		
	\end{subfigure}
	\centering
	\begin{subfigure}{0.22\linewidth}
		\centering
		\includegraphics[width=38mm, height=46mm]{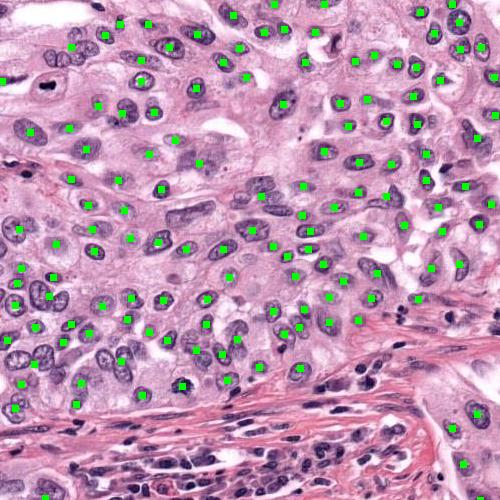}
		\caption{Ground Truth2}
		
	\end{subfigure}
    \begin{subfigure}{0.22\linewidth}
		\centering
		\includegraphics[width=38mm, height=46mm]{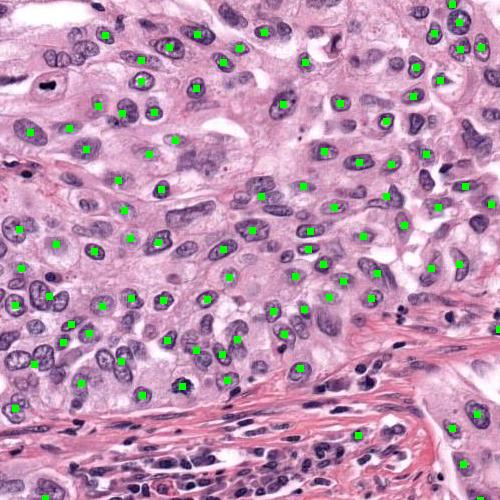}
		\caption{Prediction2}
		
	\end{subfigure}
    \vspace{-5pt}
	\caption{Visual comparison of ground truth and UAM predictions. Tumor cells are highlighted in green.}
	\label{subfig:vis5}
\end{figure*}

\begin{figure*}[htbp]
\vspace{-9pt}
	\centering
	\begin{subfigure}{0.22\linewidth}
		\centering
		\includegraphics[width=38mm, height=46mm]{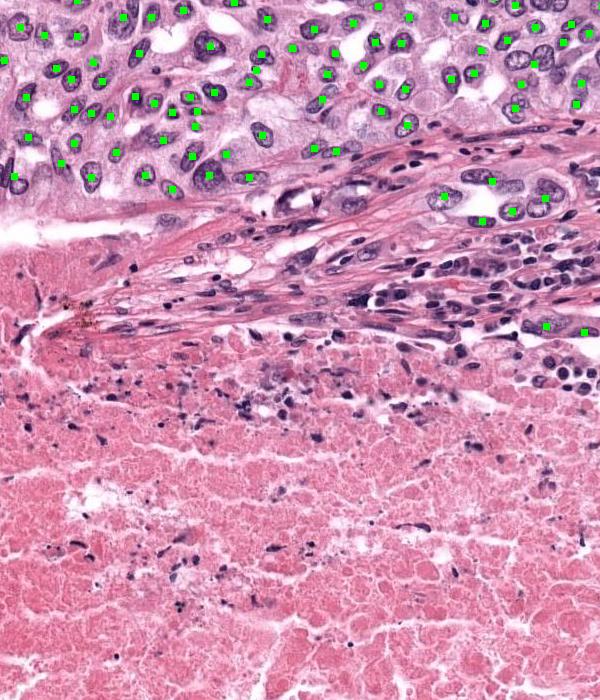}
		\caption{Ground Truth1}
		
	\end{subfigure}
	\centering
	\begin{subfigure}{0.22\linewidth}
		\centering
		\includegraphics[width=38mm, height=46mm]{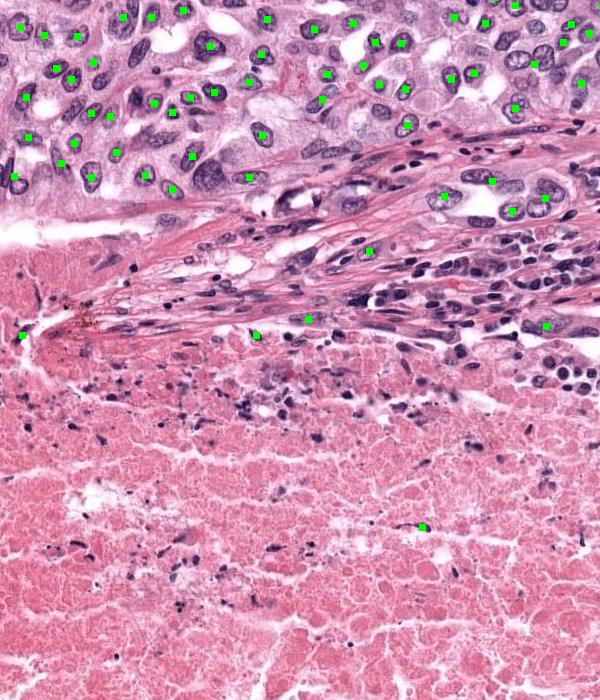}
		\caption{Prediction1}
		
	\end{subfigure}
	\centering
	\begin{subfigure}{0.22\linewidth}
		\centering
		\includegraphics[width=38mm, height=46mm]{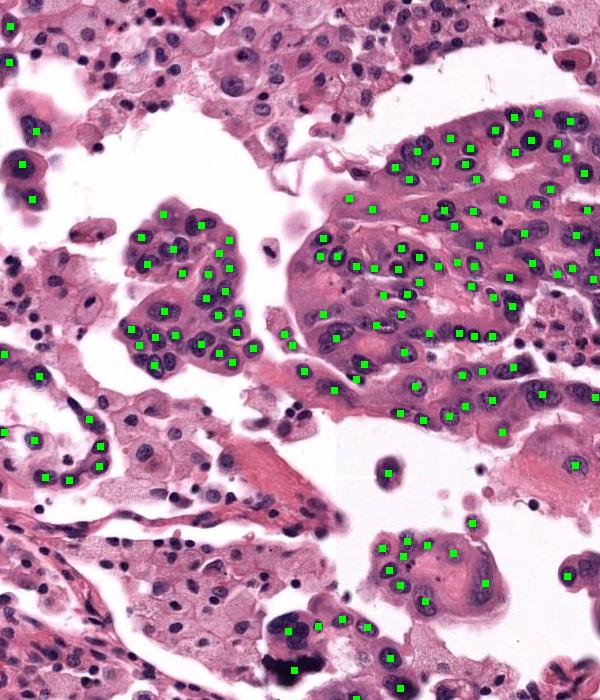}
		\caption{Ground Truth2}
		
	\end{subfigure}
    \begin{subfigure}{0.22\linewidth}
		\centering
		\includegraphics[width=38mm, height=46mm]{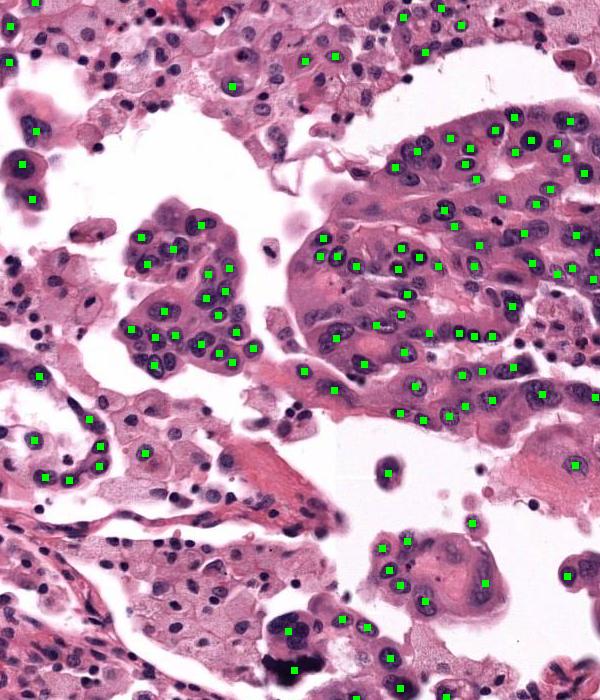}
		\caption{Prediction2}
		
	\end{subfigure}
    \vspace{-5pt}
	\caption{Visual comparison of ground truth and UAM predictions. Tumor cells are highlighted in green.}
	\label{subfig:vis6}
\end{figure*}

We provide a visualization of the Ground Truth, BiomedParse-generated, and UAM-generated masks in Figure \ref{subfig:vis2}. The multimodal UAM model demonstrates the capability to provide a more accurate mask, which we attribute to its effectiveness in leveraging the cell radiomics information from each image. This finding underscores the significant advantage and effectiveness of the multimodal UAM architecture. More visualization results are shown in Figures~\ref{subfig:vis3},~\ref{subfig:vis4},~\ref{subfig:vis5},~\ref{subfig:vis6}, indicating the effectiveness of UAM that highlights tumor cells on H\&E slides based on radiomics data, facilitating pathologist interpretation. 

\end{document}